\documentclass[11pt,a4paper]{article}
\pdfoutput=1
\usepackage[hyperref]{style/emnlp2020}
\usepackage{times}
\usepackage{latexsym}
\usepackage{url}
\usepackage{booktabs}
\usepackage{tikz}
\usepackage{subfigure}
\usepackage[utf8]{inputenc}
\usepackage[T1]{fontenc}
\usepackage{anyfontsize}
\usepackage{pifont}% http://ctan.org/pkg/pifont
\usepackage{csquotes}

\aclfinalcopy % Uncomment this line for the final submission
\def\aclpaperid{527} %  Enter the acl Paper ID here
\newcommand{\dsurl}{\href{http://curiosity.pedro.ai}{curiosity.pedro.ai}}
\newif\ifcomment
\commentfalse
\usepackage{framed}
\usepackage{mdwlist}
\usepackage{latexsym}
\usepackage{xcolor}
\usepackage{nicefrac}
\usepackage{booktabs}
\usepackage{amsfonts}
\usepackage[T1]{fontenc}
\usepackage{bold-extra}
\usepackage{amsmath}
\usepackage{dsfont}
\usepackage{amssymb}
\usepackage{bm}
\usepackage{graphicx}
\usepackage{mathtools}
\usepackage{microtype}
\usepackage{multirow}
\usepackage{multicol}
\usepackage{latexsym,comment}

\newcommand{\gem}[1]{\mbox{\textsc{gem}}}
\newcommand{\abr}[1]{\textsc{#1}}

\newcommand{\g}{\, | \,}

\newcommand{\emaillink}[1]{ {\small \href{mailto://#1}{\texttt{#1}}}}

\newcommand{\hidetext}[1]{}
\newcommand{\ignore}[1]{}

\ifcomment
    \newcommand{\pinaforecomment}[3]{\colorbox{#1}{\parbox{.8\linewidth}{#2: #3}}}
\else
    \newcommand{\pinaforecomment}[3]{}
\fi

\newcommand{\smallurl}[1]{ \begin{tiny}\url{#1}\end{tiny}}

\definecolor{lightblue}{HTML}{3cc7ea}
\definecolor{CUgold}{HTML}{CFB87C}
\definecolor{grey}{rgb}{0.95,0.95,0.95}
\definecolor{ceil}{rgb}{0.57, 0.63, 0.81}
\definecolor{vyellow}{HTML}{d1c73b}
\newcommand*{\TakeFourierOrnament}[1]{{%
            \fontencoding{U}\fontfamily{futs}\selectfont\char#1}}
\newcommand*{\danger}{\TakeFourierOrnament{66}}
\newcommand\cmark {\textcolor{green}{\ding{52}}}
\newcommand\xmark {\textcolor{red}{\ding{55}}}
\newcommand\dmark {\textcolor{vyellow}{\danger}}

\newcommand{\coqa}{\textsc{c}{\small o}\textsc{q}{\small a}}
\newcommand{\quac}{\textsc{q}{\small u}\textsc{ac}}
\newcommand{\glove}{\abr{gl}{\small o}\abr{ve}}
\newcommand{\opendialkg}{\abr{o}{\small pen}\abr{d}{\small ial}\abr{kg}}
\newcommand{\mrr}{\abr{mrr}}
\newcommand{\fone}{\abr{f}$_1$}
\newcommand{\tfidf}{\abr{tf-idf}}
\newcommand{\hre}{\abr{hre}}
\newcommand{\parlai}{\abr{p}{\small arl}\abr{ai}}
\newcommand{\wow}{\abr{w}{\small o}\abr{w}}

\newcommand{\woz}{\abr{w}{\small o}\abr{z}}
\newcommand{\allennlp}{\abr{A}{\small llen}\abr{nlp}}
\newcommand{\saac}{\abr{s}{\small aa}\abr{c}}
\newcommand{\cast}{\abr{ca}{\small s}\abr{t}}
\newcommand{\charm}{\abr{charm}}

\usepackage{graphicx}
\usepackage{multicol}
\usepackage{subfigure}
\usepackage{booktabs}
\usepackage{algorithm}
\usepackage{algorithmic}
\usepackage{color}
\usepackage{amsmath}
\usepackage{amssymb}
\usepackage{amsfonts}
\usepackage{multirow, varwidth}
\usepackage{stfloats}
\usepackage{verbatim}
\usepackage[normalem]{ulem} % [normalem] prevents the package from changing the default behavior of `\emph` to underline.
\makeatletter
\newif\if@restonecol
\makeatother
\usepackage{xr}
\usepackage[amsmath,thmmarks]{ntheorem}

\DeclareRobustCommand\onedot{\futurelet\@let@token\@onedot}
\def\onedot{. }

\newcommand{\rover}{Curiosity}
\newcommand{\bert}{\textsc{bert}}
\newcommand{\mtbert}{\abr{e2e} \bert{}}

\newcommand{\topic}[1]{\underline{#1}}
\newcommand{\aspect}[1]{\textit{#1}}
\newcommand{\entity}[1]{\dashuline{#1}}
\newcommand{\nicebox}[1]{\tikz\node[draw=white!40!black,inner sep=1pt,line width=0.3mm,rounded corners=0.1cm]{#1};}

\newcommand{\nfactsfull}{0}
\newcommand{\ndialogsfull}{0}
\newcommand{\nutterfull}{0}
\newcommand{\ntopicsfull}{0}
\newcommand{\nshownfull}{0}
\newcommand{\nusedfull}{0}
\newcommand{\nlikedfull}{0}

\newcommand{\ndialogstrain}{0}
\newcommand{\nuttertrain}{0}
\newcommand{\ntopicstrain}{0}
\newcommand{\nshowntrain}{0}
\newcommand{\nusedtrain}{0}
\newcommand{\nlikedtrain}{0}

\newcommand{\ndialogsval}{0}
\newcommand{\nutterval}{0}
\newcommand{\ntopicsval}{0}
\newcommand{\nshownval}{0}
\newcommand{\nusedval}{0}
\newcommand{\nlikedval}{0}

\newcommand{\ndialogstest}{0}
\newcommand{\nuttertest}{0}
\newcommand{\ntopicstest}{0}
\newcommand{\nshowntest}{0}
\newcommand{\nusedtest}{0}
\newcommand{\nlikedtest}{0}

\newcommand{\ndialogstestzero}{0}
\newcommand{\nuttertestzero}{0}
\newcommand{\ntopicstestzero}{0}
\newcommand{\nshowntestzero}{0}
\newcommand{\nusedtestzero}{0}
\newcommand{\nlikedtestzero}{0}

\newcommand{\kripscore}{0.00}

\IfFileExists{2020_emnlp_curiosity/auto_fig/dataset-stats-full.tex}{
\renewcommand{\nfactsfull}{93,845}
\renewcommand{\ndialogsfull}{14,048}
\renewcommand{\nutterfull}{181,068}
\renewcommand{\ntopicsfull}{361}
\renewcommand{\nshownfull}{76,120}
\renewcommand{\nusedfull}{27,486}
\renewcommand{\nlikedfull}{57,607}}{}
\IfFileExists{2020_emnlp_curiosity/auto_fig/dataset-stats-train.tex}{

\renewcommand{\ndialogstrain}{10,287}
\renewcommand{\nuttertrain}{131,394}
\renewcommand{\ntopicstrain}{331}
\renewcommand{\nshowntrain}{66,913}
\renewcommand{\nusedtrain}{21,669}
\renewcommand{\nlikedtrain}{41,015}}{}
\IfFileExists{2020_emnlp_curiosity/auto_fig/dataset-stats-val.tex}{

\renewcommand{\ndialogsval}{1,287}
\renewcommand{\nutterval}{17,186}
\renewcommand{\ntopicsval}{318}
\renewcommand{\nshownval}{29,785}
\renewcommand{\nusedval}{4,950}
\renewcommand{\nlikedval}{5,928}}{}
\IfFileExists{2020_emnlp_curiosity/auto_fig/dataset-stats-test.tex}{

\renewcommand{\ndialogstest}{1,287}
\renewcommand{\nuttertest}{17,187}
\renewcommand{\ntopicstest}{316}
\renewcommand{\nshowntest}{30,162}
\renewcommand{\nusedtest}{4,952}
\renewcommand{\nlikedtest}{5,846}}{}
\IfFileExists{2020_emnlp_curiosity/auto_fig/dataset-stats-testzero.tex}{

\renewcommand{\ndialogstestzero}{1,187}
\renewcommand{\nuttertestzero}{15,301}
\renewcommand{\ntopicstestzero}{30}
\renewcommand{\nshowntestzero}{6,043}
\renewcommand{\nusedtestzero}{2,290}
\renewcommand{\nlikedtestzero}{4,818}}{}
\IfFileExists{2020_emnlp_curiosity/auto_fig/krip-score.tex}{\renewcommand{\kripscore}{0.834}}{}
\IfFileExists{2020_emnlp_curiosity/auto_fig/student-prefs.tex}{

}{}
\graphicspath{{2020_emnlp_curiosity/figures/}{2020_emnlp_curiosity/commit_auto_fig/}{2020_emnlp_curiosity/auto_fig/}}

\title{Information Seeking in the Spirit of Learning:\\A Dataset for Conversational Curiosity}

\author{
    Pedro Rodriguez\textsuperscript{\thanks{$^\star$Work done while interning at Facebook.}} \\
    Computer Science\\
    University of Maryland\\
    \emaillink{pedro@cs.umd.edu}
    \And
    Paul Crook\\
    Facebook \\
    \emaillink{pacrook@fb.com}
    \AND
    Seungwhan Moon \\
    Facebook \\
    \emaillink{shanemoon@fb.com}
    \And
    Zhiguang Wang \\
    Facebook \\
    \emaillink{zgwang@fb.com}
}

\date{}

\begin{document}
\maketitle
\setlength{\abovedisplayskip}{3pt}
\setlength{\belowdisplayskip}{3pt}

\begin{abstract}
  Open-ended human learning and information-seeking are increasingly mediated by digital assistants.
However, such systems often ignore the user's pre-existing knowledge.
Assuming a correlation between engagement and user responses such as ``liking'' messages or asking followup questions, we design a Wizard-of-Oz dialog task that tests the hypothesis that engagement increases when users are presented with facts related to what they know.
Through crowd-sourcing of this experiment, we collect and release 14K dialogs (181K utterances) where users and assistants converse about geographic topics like geopolitical entities and locations.
This dataset is annotated with pre-existing user knowledge, message-level dialog acts, grounding to Wikipedia, and user reactions to messages.
Responses using a user's prior knowledge increase engagement.
We incorporate this knowledge into a multi-task model that reproduces human assistant policies and improves over a \bert{} content model by 13 mean reciprocal rank points.

\end{abstract}

\section{Introduction}
\label{sec:intro}

Conversational agents such as Alexa, Siri, and Google Assistant should help users discover, learn, and retain novel factual information.
More generally, systems for conversational information-seeking should help users develop their information need, be mixed-initiative, incorporate user memory, and reason about the utility of retrieved information as a combined set~\citep{Radlinski2017ATF}.
We focus on a curiosity-driven, information-seeking scenario where a user starts a conversation with an assistant by asking an open-ended question and then drills down into interest areas (Figure~\ref{fig:example}).
\begin{figure}[t]
\centering
\begingroup
\addtolength\leftmargini{-5mm}
\begin{itemize}
    \setlength\itemsep{-1.25mm}
          \it
          \fontsize{10}{12}\selectfont
    \item[] U: <assistant wake-word>, tell me about Tahiti.
    \item[] A: It's the largest island in French Polynesia, near the center of the Pacific
    \item[] U: What is its history with France?
\end{itemize}
\endgroup
\caption{
    An example of information-seeking dialog that the \rover{} dataset aims to support.
    Assistants should answer user questions \emph{and} convey information that inspires meaningful followup questions.
}
\label{fig:example}
\end{figure}

In this setting, what policies should assistants pursue to maintain the user's interest in the topic?
Theories of human learning, such as Vygotsky's zone of proximal development, propose that learning novel information should be rooted in pre-existing knowledge and skills of the learner~\citep{chaiklin-03}.
Considering this, a good policy may give general information about Tahiti; a better policy would select information related to the user's knowledge (e.g., familiarity with France).
We hypothesize that engagement is correlated with policies that integrate a user's pre-existing knowledge, and test this through a large-scale, Wizard-of-Oz~(\woz{}) style collection~\citep{Kelley1984AnID,Wen2016ANE} that captures assistant policies, user reactions, and topically relevant entities that the user knows about.
The \rover{} dataset has \ndialogsfull{} English dialogs annotated with sentence-level knowledge grounding, the user's prior knowledge, dialog acts per message, and binary ratings per message.\footnote{
    Dataset and code at \dsurl{}.
}

\begin{figure}[t]
    \centering
    \includegraphics[width=\linewidth]{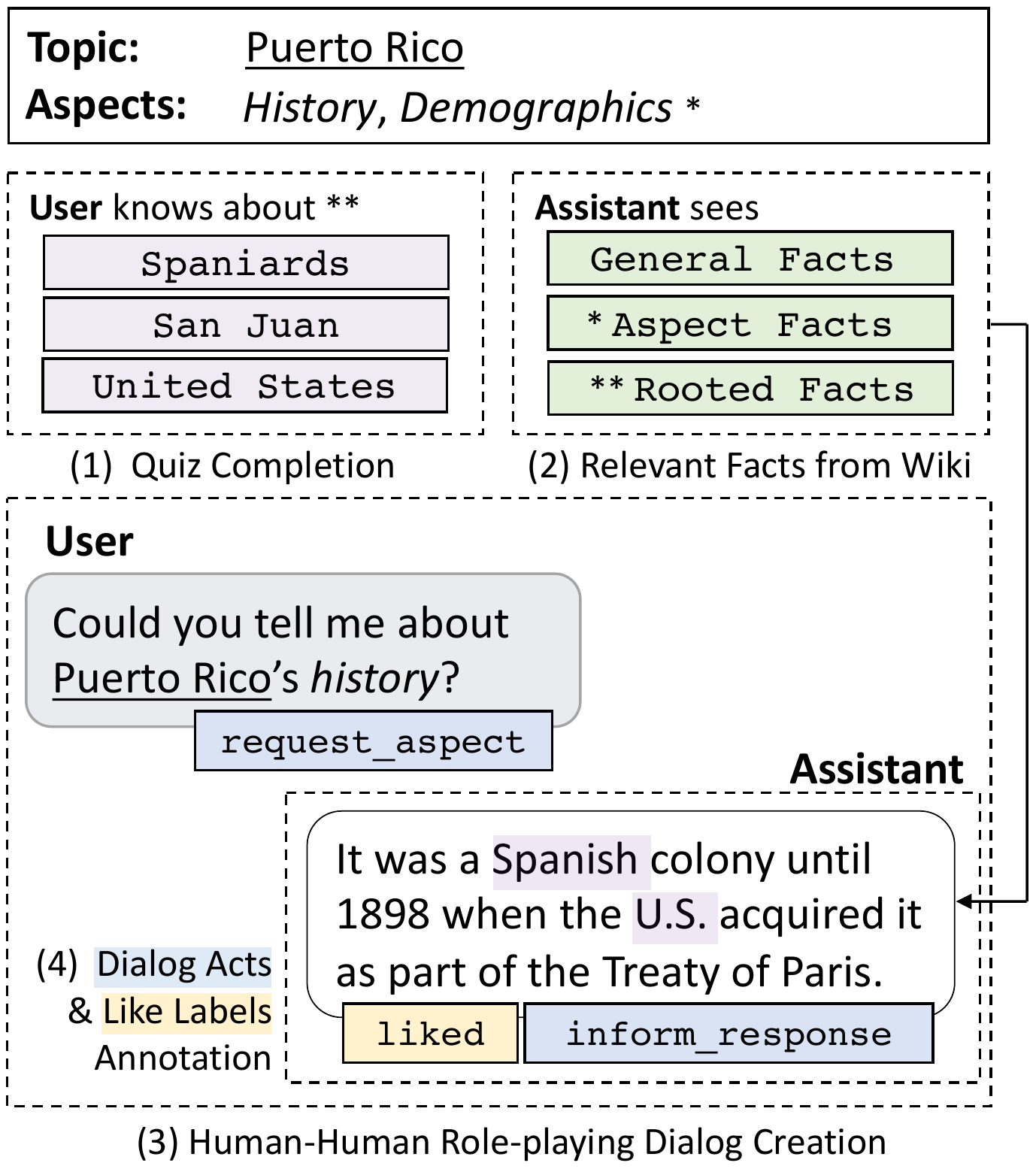}
    \vspace{-16pt}
    \caption{
        We sample pre-existing knowledge by asking users to indicate which \topic{topically} related entities they already \emph{know}.
        The assistant paraphrases facts related to either known entities (rooted facts), an aspect (aspect facts), or the topic generally (general facts).
        The user expresses engagement through a like button.
        Dialog acts are annotated in a separate crowd-source task.
    }
    \vspace{-16pt}
    \label{fig:ex-dia}
\end{figure}

In our dialog task (Figure~\ref{fig:ex-dia}), one worker takes the role of a curious user learning about a geographic entity and the other of a digital assistant with access to Wikipedia facts (Section~\ref{sec:data}).
At the start of each dialog, the user is assigned an entity as their \topic{topic} (e.g., \topic{Puerto Rico}) along with two \aspect{aspects} (e.g., \aspect{history} and \aspect{demographics}) to investigate.
Beforehand, we show the user a list of \entity{entities} related to the \topic{topic}, and they mark which they know; these entities are a sample of their pre-existing knowledge.
The user engages in open-ended discovery while the assistant simultaneously answers the user's questions and proactively introducing facts likely to prompt followup questions.
%For example, if the assistant knew of a user's familiarity with \entity{astronomy} when providing information about \topic{Puerto Rico}, then the user is more likely to engage with and remember facts about the \entity{Arecibo Observatory}.

Section~\ref{sec:analysis} uses dialog act annotations combined with explicit and implicit user feedback to compare assistants' content selection and presentation policies.
For example, in interactions where the user asks a question and the assistant paraphrases a fact, how often does the user ask a followup question versus trail off in disinterest?
Most datasets (Section~\ref{sec:rel}) do not have enough annotations to answer these questions: it requires message-level dialog act annotations and feedback signals.
We compare three assistant policies: using a fact with a rooted entity, a fact from the user's aspect, or a generic fact about the topic.
The policies are compared through user ``likes'' of assistant messages and by the dialog act of their subsequent message~(e.g., did they ask a specific followup or change topic).

In Section~\ref{sec:method}, we design models that predict the policies used by the assistant: what type of message to send and which fact to use~(if any).
All models are trained jointly with a multi-task objective function.
We compare an end-to-end \bert{}~\citep{Devlin2018BERTPO} model to our task-specific Hierarchical Recurrent Encoder model~\citep{Serban2015BuildingED} and show that our model improves over the baseline.

In summary, we make three main contributions: (1) we design an experiment to test the efficacy of personalizing conversational information systems through a user's prior knowledge,  (2) introduce the \rover{} dataset---the first dialog dataset combining sentence-level knowledge groundings, per message ratings, \emph{and} per message dialog act annotations, allowing for robust and fine-grained structural learning of dialog policies for similar applications, and (3) design a multi-task model that incorporates the user's prior knowledge and improves over a natural \bert{} baseline.

\section{Building the \rover{} Dataset}
\label{sec:data}

This section describes the construction of the \rover{} dataset.
Dialog topics consist of prominent world geographic entities.
The \emph{worldwide} spread of entities makes each novel to most users, the consistent topic type makes starting dialogs easier, and their rich histories, demographics, and economics add topical diversity.
For example, most people are only vaguely familiar with the history of \topic{Puerto Rico}, but most know about related concepts such as the \entity{United States} or \entity{Hurricane Maria}.
Section~\ref{sec:geo} describes how we select geographic topics, aspects, and derive a set of facts to ground against.
We collected the dataset in two steps: (1) collecting dialogs with a custom interface (Section~\ref{sec:ints}) and (2) after-the-fact dialog act annotation (Section~\ref{sec:da}).
Sample dialogs from \rover{} are in Appendix~\ref{apx:samples}.

\subsection{Geographic Topics, Aspects, and Facts}
\label{sec:geo}
We select \ntopicsfull{} geographic pages from Wikipedia that have separate geography and history pages (e.g., \topic{Puerto Rico}, \entity{Geography of Puerto Rico}, and \entity{History of Puerto Rico}).\footnote{
    The existence of these pages implies that the topic has ample historical and geographical knowledge to draw from.
}
We use sentences from each page to build a set of \nfactsfull{} facts.
We run an entity linker over the content~\citep{gupta-etal-2017-entity} and index each fact by its source page (\topic{topic}), source section (\aspect{aspect}), and mentioned entities.
Finally, we fit a \textsc{tf-idf} text matcher~\citep{rajaraman_ullman_2011} with Scikit-Learn~\citep{scikit-learn}.
While conversing, assistants are shown facts filtered by topic, aspect, or mentioned entities, that are ranked by textual similarity to the dialog.

\subsection{User and Assistant Dialog Interfaces}
\label{sec:ints}

To collect dialogs, we build user and assistant interfaces for annotators.
The user's interface samples their prior knowledge of a topic, captures which assistant messages interest them, and manages the dialog context.
The assistant's interface provides contextually relevant facts.
Appendix~\ref{apx:int-photos} has screenshots and details of each interface component.

\paragraph{Sampling User's Prior Knowledge}
When deployed, digital assistants can draw from prior interactions~\citep{Ram2018ConversationalAT} to estimate what a user knows.
However, since we do not have these prior interactions, we collect information about what users know.
Instead of exhaustively asking about every entity related to the topic, we sample this knowledge.
Before the dialog begins, we show the user fifteen related entities that range from commonplace to obscure (\entity{United States} versus \entity{Ta\'ino}).
Users mark the entities they could (1) locate on a map or (2) summarize succinctly in one sentence.

\paragraph{Like Button for User Interest}
As part of our collection, we aimed to determine what fact-grounded utterances users found interesting.
Users ``liked'' the assistant's message if they found it ``interesting, informative, and relevant to their topic.''

\paragraph{Assistant's Topic Summary and Fact Bank}
The worldwide spread of \rover{}'s entities makes them unfamiliar to most crowd-workers, including the assistants.
So that the assistant can still engage the user, the assistant interface provides contextually relevant information.
First, the interface shows a topic summary from Wikipedia.
Second, the assistant paraphrases facts from a contextually updated fact bank (box 2 in Figure~\ref{fig:ex-dia}).
To reduce information overload, we use simplified topic descriptions from SimpleWikipedia and show a maximum of nine facts at a time.\footnote{
    If a description exists in \href{https://simple.wikipedia.org/}{simple.wikipedia.org}, we use that; otherwise, we use the description from \href{https://en.wikipedia.org/}{en.wikipedia.org}.}
We encourage assistants to ``stimulate user interest and relate information to things they already know or have expressed interest in.''
Assistants are instructed to select relevant facts, click the ``use'' button, and paraphrase the content into their next utterance.

Like~\citet{dinan2019wizard}, the fact bank shows facts to the assistant using \tfidf{} textual similarity to recent dialog turns but differs by incorporating the user's prior knowledge.
We show the assistant nine facts: three facts that mention an entity familiar to the user (rooted facts), three facts from their assigned aspects (aspect facts), and three from anywhere on the page (general facts).
By construction, rooted facts overlap with the exclusive categories of aspect and general facts.
For each category, we find the nine highest \abr{tf-idf} scoring facts and then randomize their order.
To avoid biasing the assistant, we do not inform them about the user's known entities or distinguish between types of facts.

\subsection{Dialog Act Annotation}
\label{sec:da}

Inducing structure on conversations through dialog acts is helpful for analysis and downstream models~\citep{tanaka-etal-2019-dialogue}.
We introduce structure---beyond knowledge groundings---into \rover{} by annotating dialog acts for each message.

In a separate collection, we annotate all utterances with dialogs acts using a custom interface~(Appendix~\ref{apx:acts}).
The annotation schema is based on \abr{iso} $24617$-2~\citep{Bunt2010TowardsAI,Bunt2012ISO2A} with customized sub-categories for our scenario.
Table~\ref{tbl:acts} shows our schema, descriptions, and examples.

% feedback_ask 36
% feedback_negative 176
% feedback_positive 26946
% inform_related 6981
% inform_response 59269
% inform_unrelated 557
% offer_accept 1727
% offer_aspect 1440
% offer_decline 405
% offer_followup 63
% offer_other 1619
% offer_topic 91
% request_aspect 41701
% request_followup 4463
% request_other 10077
% request_topic 10789
% shift_aspect 201
% social_apology 27
% social_goodbye 90
% social_greeting 200
% social_thanking 399
\begin{table*}[ht]
    \centering
    \small
    \begin{tabular}{l r l l}
        \toprule
        \textbf{Dialog Act} & \textbf{Count} & \textbf{Description}                   & \textbf{Example}                                     \\
        \midrule
        request topic       & $10,789$       & A request primarily about the topic.   & I'd like to know about \topic{Puerto Rico}.          \\
        request aspect      & $41,701$       & A request primarily about an aspect.   & Could you tell me about its \aspect{history}?        \\
        request followup    & $4,463$        & A request about mentioned concept.     & Do you know more about the \entity{Ta\'inos}?        \\
        request other       & $10,077$       & Requests on unmentioned concepts.      & What is there to know about cuisine?                 \\
        \midrule
        inform response     & $59,269$       & Directly answer an info request.       & \entity{Ta\'inos} were caribbean indigenous.         \\
        inform related      & $6,981$        & Not a direct answer, but related info. & I do not know, but\ldots                             \\
        inform unrelated    & $557$          & Does not answer question, not related. & Politics is tiring!                                  \\
        \midrule
        feedback positive   & $26,946$       & Provide positive feedback              & Thats quite interesting!                             \\
        feedback negative   & $176$          & Provide negative feedback              & Thats pretty boring.                                 \\
        feedback ask        & $36$           & Ask for feedback                       & Do you find \textless~info~\textgreater~interesting? \\
        \midrule
        offer topic         & $91$           & Offer to discuss topic                 & Want to learn about \topic{Puerto Rico}?             \\
        offer aspect        & $1,440$        & Offer to discuss aspect                & How about more on its \aspect{demographics}?         \\
        offer followup      & $63$           & Offer to discuss mentioned concept.    & I could say more about the \entity{Spanish}.         \\
        offer other         & $1,619$        & Offer to discuss unmentioned concept.  & How about I tell you about its exports.              \\
        offer accept        & $1,727$        & Accept offer of information.           & I'd love to learn about its \topic{history}.         \\
        offer decline       & $405$          & Decline offer of information           & Sorry, I'm not interested in that.                   \\
        \bottomrule
    \end{tabular}
    \caption{
        Counts, abbreviated descriptions and examples of the dataset's dialog acts.
    }
    \label{tbl:acts}
\end{table*}

\subsection{Data Quality}
\label{sec:collection}
We crowd-sourced conversations in two phases using \parlai{}~\citep{miller2017parlai}.
In the first, pilot studies collect feedback from individual workers.
Based on feedback, we create task guidelines, sample dialogs, a \abr{faq}, tutorial videos, and qualification tests.
These materials were used to train and qualify crowd-workers for the second phase.
During the second, we monitor the interface usage and removed workers that ignored instructions.

\begin{table}
    \small
    \centering
    \begin{tabular}{ c c c }
                             & Annotator 1 & Annotator 2 \\
        \toprule
        Utterance 1, Label A & Yes         & No          \\
        Utterance 1, Label B & Yes         & No          \\
        Utterance 2, Label A & Yes         & Yes         \\
        Utterance 2, Label B & Yes         & Yes         \\
        \bottomrule
    \end{tabular}
    \caption{
        Consider a task where each utterance has labels A and B.
        In the single-label version, each utterance is labeled as either A or B.
        The table shows the outcome of converting the multi-label version to single-label by creating a row for each example--label combination.
        Cell values are binary indicators.
    }
    \label{table:krip-multi}
\end{table}

Using Krippendorff's $\alpha$~\citep{kripp2004}, we validate the quality of dialog act annotations.
Dialog acts are multi-class and multi-label: a message can have none, one, or multiple dialog acts (e.g., positive feedback and followup).
However, Krippendorff's $\alpha$ is computed for single-label tasks from a table where rows represent examples, columns represent annotators, and cells indicate the singular class label.
We convert our multi-label problem to a single label problem by making each combination of example and label class a row in the table (Table~\ref{table:krip-multi}).
Since there are few dialog acts per utterance, most annotations agree; however, since Krippendorff's $\alpha$ focuses on disagreement, it is appropriate for this scenario.
Using a separate annotation interface (Appendix~\ref{apx:acts}), we doubly annotate 4,408 dialogs and the agreement score \kripscore{} is higher than the 0.8 threshold recommended by \citet{kripp2004}.
Next, we analyze the annotated dialogs and introduce our model.

\section{Dataset Analysis}
\label{sec:analysis}

This section shows statistics of the \rover{} dataset and that users prefer aspect-specific, rooted facts.

\subsection{Dataset Statistics}
\label{sec:stats}
Table~\ref{tbl:stats} shows the basic statistics of the \rover{} dataset.
In total, our dataset contains \ndialogsfull{} dialogs with \nutterfull{} utterances.
The fact database contains \nfactsfull{} facts; of those, \nshownfull{} ($81\%$) were shown to the assistants and \nusedfull{} ($29\%$) were used in at least one message.
We randomly split dialogs into training, validation, and testing folds.

\begin{table}[t]
    \centering
    \scalebox{0.75}{
        \begin{tabular}{l r r r r r}
            \toprule
            \textbf{Metric (\# of)} & \textbf{Total}  & \textbf{Train}   & \textbf{Val}   & \textbf{Test}   & \textbf{Zero}       \\
            \midrule
            Dialogues               & \ndialogsfull{} & \ndialogstrain{} & \ndialogsval{} & \ndialogstest{} & \ndialogstestzero{} \\
            Utterances              & \nutterfull{}   & \nuttertrain{}   & \nutterval{}   & \nuttertest{}   & \nuttertestzero{}   \\
            Likes                   & \nlikedfull{}   & \nlikedtrain{}   & \nlikedval{}   & \nlikedtest{}   & \nlikedtestzero{}   \\
            Topics                  & \ntopicsfull{}  & \ntopicstrain{}  & \ntopicsval{}  & \ntopicstest{}  & \ntopicstestzero{}  \\
            Facts Total             & \nfactsfull{}   & NA               & NA             & NA              & NA                  \\
            Facts Shown             & \nshownfull{}   & \nshowntrain{}   & \nshownval{}   & \nshowntest{}   & \nshowntestzero{}   \\
            Facts Used              & \nusedfull{}    & \nusedtrain{}    & \nusedval{}    & \nusedtest{}    & \nusedtestzero{}    \\
            \bottomrule
        \end{tabular}
    }
    \caption{
        \rover{} has \ndialogsfull{} dialogs. On average, dialogs have $12.9$ utterances.
        $60\%$ of the assistants' 90,534 utterances were liked.
    }
    \label{tbl:stats}
\end{table}

\subsection{What Facts do User Prefer?}

\begin{figure}[t]
    \centering
    \includegraphics[width=\linewidth]{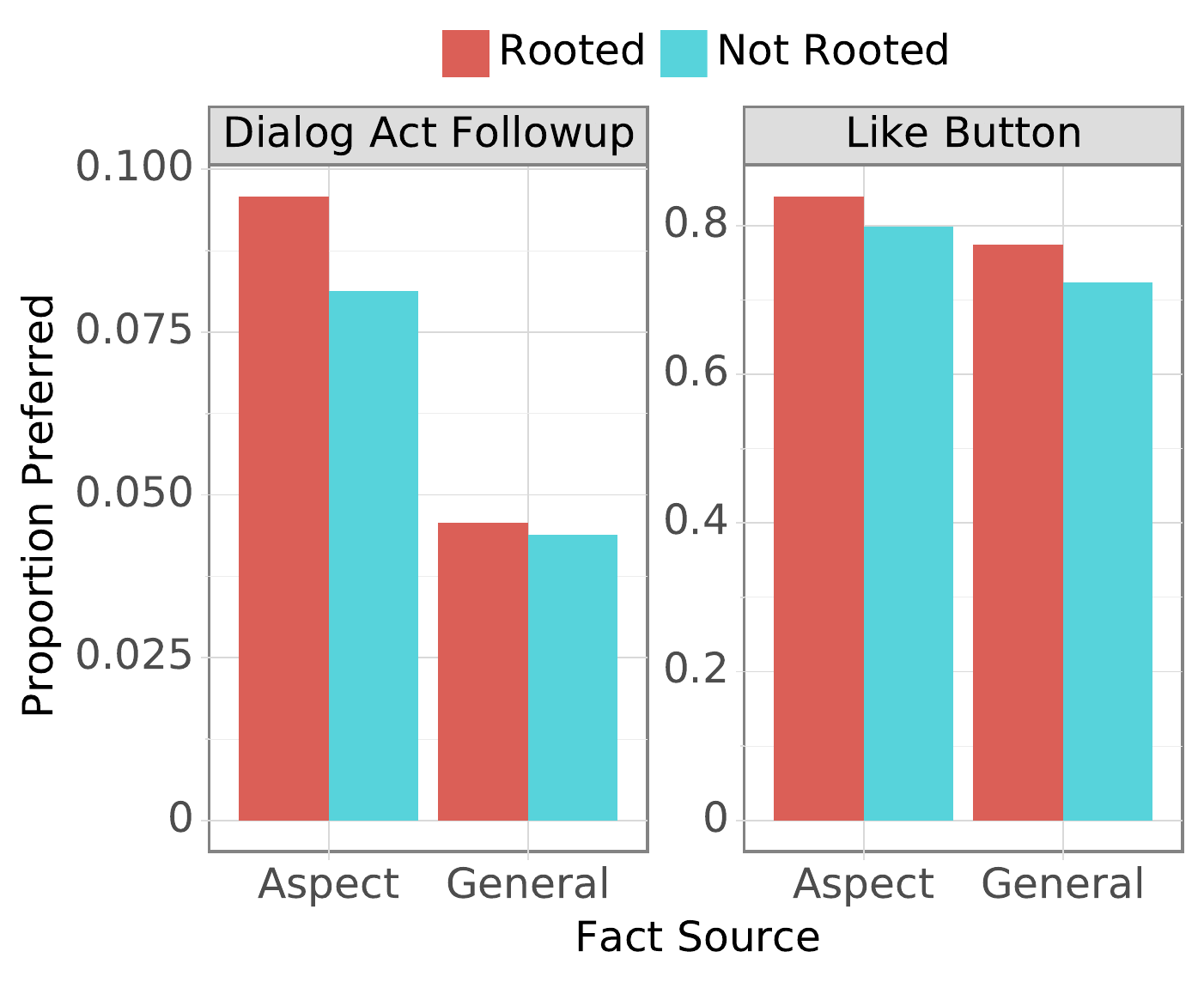}
    \caption{
        User engagement is measured by dialog act followups (left) and like button usage (right).
        We compare reactions to messages that use a fact mentioning an entity the user knew about (rooted) and whether the fact is general or aspect-specific.
        Pairwise differences are statistically significant ($99\%+$) with a two proportion z-test \emph{except} for dialog act followups between rooted and non-rooted general facts.
        Overall, users prefer on-aspect, rooted facts.
    }
    \vspace{-12pt}
    \label{fig:prefs}
\end{figure}

In Section~\ref{sec:intro}, we hypothesized that when assistants use facts that mention previously known entities (rooted facts), users will be more likely to engage.
In our data collection, we incorporate two mechanisms to test this hypothesis.
The first mechanism is explicit: we directly ask users---through a like button---to indicate what messages they preferred.
The second mechanism is implicit and derived by mining dialogs for specific sequences of dialog acts that suggest engagement with the content.
For each of these mechanisms, we compute the likelihood $P(\text{Prefer}\g\text{Fact Source})$ of a user preferring utterances grounded to each fact source (Rooted, Aspect, or General).
Figure~\ref{fig:prefs} shows this likelihood and indicates that users prefer: (1) facts relevant to aspects versus general ones and (2) rooted facts in three of four scenarios.

\subsubsection{Likes for Explicit Preference Elicitation}
Explicit preference is computed directly from like button usage and shown on the right panel of Figure~\ref{fig:prefs}.
Overall, users liked $60\%$ of messages, and they prefer on-aspect, rooted facts.

\subsubsection{Mining Acts for Implicit Preferences}
When users ask specific followup questions---as opposed to generic ones---about an assistant's fact, it shows that the user implicitly prefers these kinds of messages.
For example, asking about an entity like the Ta\'inos is more specific than asking about history and therefore indicates engagement.
We identify these interactions by mining for pairs of assistant-user messages where the assistant uses a fact and their message is labeled with an ``inform'' dialog act.
With these, we compute the likelihood
$$P(\text{Outcome}=\text{request\ followup}\g\text{Fact Source})$$
that the user's message has the ``request followup'' dialog act given the source.
Similarly to likes, users engage more with aspect-oriented and rooted facts.

\subsubsection{Paraphrase Analysis}
\label{sec:para-analysis}
Although our work does not include a paraphrase model, we manually analyze a random sample of two hundred and fifty assistant messages where facts were used. % where annotators indicated they used a fact
Of these messages, $51\%$ were acceptable paraphrases, $27\%$ were verbatim copies, $12\%$ were contextualizations of near copies, and the remainder were errors such as incorrect paraphrases or did not incorporate the fact.
Appendix~\ref{apx:para} shows descriptions, counts, and random examples of each category.
This analysis estimates that about half of grounded messages have non-trivial signal for future paraphrase models to use.

\section{Models}
\label{sec:method}

\begin{figure*}[t]
    \centering
    \includegraphics[width=\linewidth]{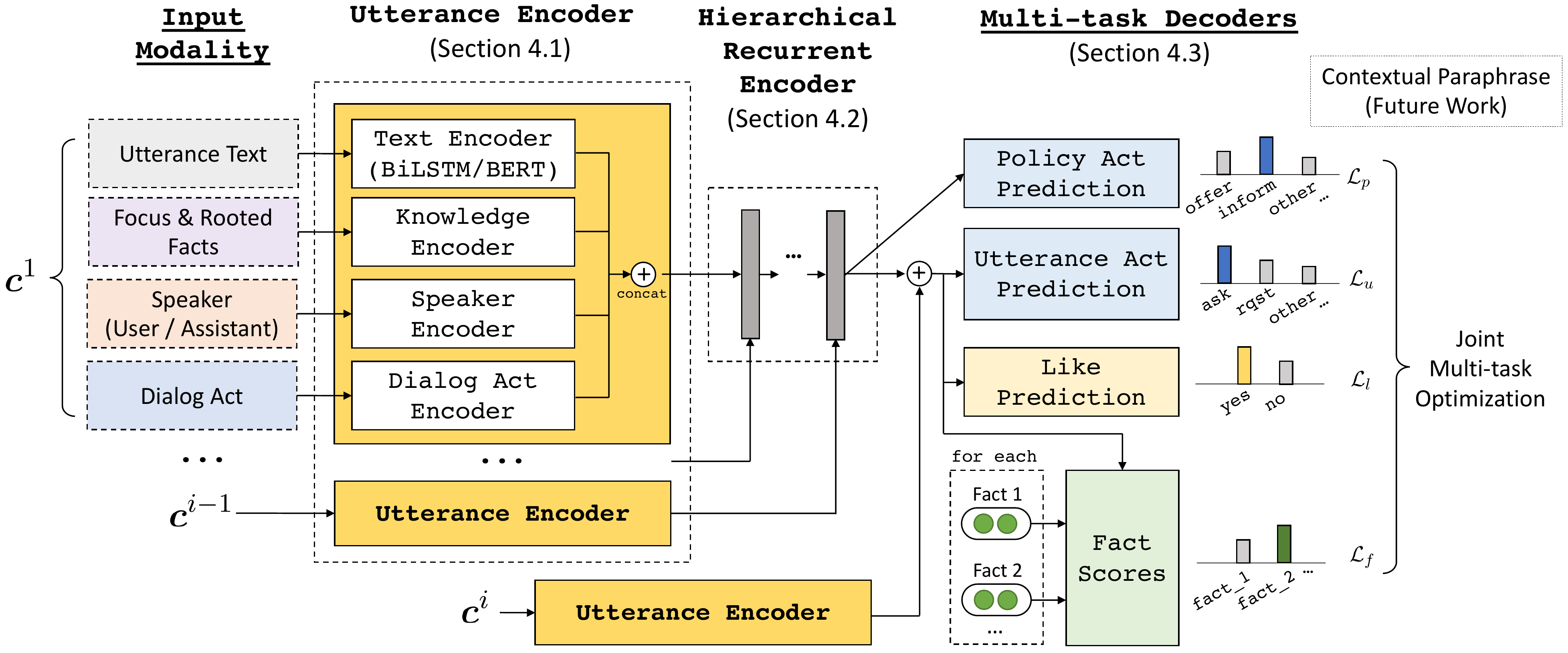}
    \caption{
        \textbf{Architecture}: \abr{charm} builds a dialog context up to $t=i-1$ to predict the current message's dialog acts~(policy prediction) and the best facts to use.
        The model uses this combined with the current utterance to classify it's dialog acts and if it will be liked.
    }
    \label{fig:model}
\end{figure*}

We design a machine learning model that predicts assistant and user actions.
We introduce a multi-task architecture for \textbf{C}uriosity that \textbf{H}ier\textbf{ar}chically \textbf{M}odels (\charm{}, Figure~\ref{fig:model}) dialogs to:
(1) predict the dialog acts of the user message~(utterance act prediction),
(2) select the best fact~(fact prediction),
(3) choose the best set of dialog acts for the next message~(policy act prediction),
and (4) predict if the assistant message will be liked~(like prediction).

\subsection{Text Representation}
\label{subsec:method:text}
\charm{} jointly encodes the text of utterances and facts with one encoder.
$E$ is a bi-directional \abr{lstm}~\citep{Sutskever2014SequenceTS} over \glove{}~\citep{pennington2014glove} word embeddings and Wikipedia2Vec~\citep{yamada2018wikipedia2vec} entity embeddings.\footnote{
    In \charm{}, \bert{} was not as effective an encoder.
}
The text $t_i^u$ of utterance $u_i$ in dialog $D$ is represented as $E(t_i^u$).
Similarly, fact~$f_j$ on turn~$i$ is represented as $E(t_{i,j}^f)$ where~$j$ indexes facts shown on that turn.

\subsection{Dialog Representation}
\label{subsec:method:dialog}
In our models, we use a hierarchical recurrent encoder (\hre{}) architecture~\citep{Sordoni2015AHR,Serban2015BuildingED} where a forward \abr{lstm} contextualizes each utterance to the full dialog.
We modify the \abr{hre} model by adding additional inputs beyond the utterance's textual representation.
First, we represent user's known entities
\begin{equation}
    \bm{k}=\text{avg}(E_{\text{entity}}(e_1),\ldots,E_{\text{entity}}(e_k)))
\end{equation}
as the average of entity embeddings.
An entity embedding also represents the topic
\begin{equation}
    \bm{t}=E_{\text{entity}}(\text{topic)}
\end{equation}
of the dialog.
Next, we create trained speaker embedding $\bm{v}_s$ for the user and $\bm{v}_t$ for the assistant.
Given the set of all dialog acts $\mathcal{A}$, each utterance has a set of dialog acts $\mathcal{A}_u\in\mathcal{P}(\mathcal{A})$ where $\mathcal{P(X)}$ denotes the set of all subsets of $\mathcal{X}$.
Finally, we use an act embedder $A$ to compute an act representation
\begin{equation}
    \bm{a}^i=\frac{1}{|\mathcal{A}_u|}\sum_{a_k\in\mathcal{A}_u} A(a_k)
\end{equation}
by averaging embeddings at each turn.
The input at each step is the concatenation
\begin{equation}
    \bm{c}^i=[E(t_i^u);\bm{a}^i;\bm{t};\bm{k};\bm{v}]
\end{equation}
of the representations for text, speaker, topic, known entities, and utterance dialog acts.\footnote{
    The speaker embedding $\bm{v}$ alternates between $\bm{v}_s$ and $\bm{v}_t$.
}
With this joint representation, the contextualized dialog representation
\begin{equation}
    \bm{h}^{i-1}=\text{LSTM}(\bm{c}^1,\ldots,\bm{c}^{i-1})
\end{equation}
is the final \abr{lstm} state and includes time step~$t=i-1$.
The dialog up to and including time $i$ is
\begin{equation}
    \bm{d}^i=[\bm{h^}{i-1};\bm{c}^{i}]
\end{equation}
which emphasizes the current utterance and makes multi-task training straightforward to implement.

\subsection{Tasks and Loss Functions}
\label{subsec:method:task}
In our model, we jointly learn to predict fact usage, user likes, utterance acts, and policy acts.
\paragraph{Fact Prediction}
For every assistant turn, the model predicts which fact(s) from
$$\{f_1,\ldots,f_k\}\in \mathcal{F}^{(i)},\mathcal{F}^{(i)}\in\mathcal{P}(\mathcal{F})$$
the assistant marked as ``used'' where $\mathcal{F}$ is the set of all facts.
We frame this task as pointwise learning to rank~\citep{Li2008LearningTR}.
A fact prediction network
\begin{equation}
    \bm{s}_{j}^{f,(i)}=\text{GELU}\left(\left[\bm{W}^f\cdot \bm{h}^{(i-1)} +\bm{b}^f;E(t_j^f)\right]\right)
\end{equation}
with parameters $\bm{W}^f$ and $\bm{b}^f$ using a Gaussian Error Linear Unit~\citep{Hendrycks2017BridgingNA} outputs salience scores for each fact.
The network does not use utterance $u_i$ since it contains signal from the choice of fact.
The predictions
\begin{equation}
    \bm{\hat{y}}^{f,(i)}_j=\text{softmax}(\bm{s}_{j}^{f,(i)})
\end{equation}
are converted to probabilities by the softmax
\begin{equation}
    \text{softmax}(\bm{q})=\frac{exp(\bm{q})}{\sum_{j=1}^k exp({\bm{q}_j})}
\end{equation}
over $k$ labels. Using this, we compute the fact loss
\begin{equation}
    \mathcal{L}_f=\frac{1}{|\mathcal{F}^{(i)}|}\sum_{i,j} \ell_{ce}(\bm{\hat{y}}^f_{i,j},\bm{y}_{i,j})
\end{equation}
where labels $\bm{y}^{f,(i)}_{j}$ indicate if fact from utterance $i$ in position $j$ was used and
\begin{equation}
    \ell_{ce}(\bm{\hat{y}},\bm{y})=\sum_{p=1}^k\bm{y}_p\log(\bm{\hat{y}}_p).
\end{equation}
is the cross entropy loss.
To mitigate class imbalance, we also scale positive classes by nine~\citep{Japkowicz2002TheCI}.

\paragraph{Policy Act and Utterance Act Prediction}
Each utterance may have multiple dialog acts so we treat policy and utterance act prediction as a multi-label task.
The goal of policy prediction is to choose the best act for the next utterance; the utterance act classifies the last message's acts.
To predict these acts, we create a policy act network
\begin{equation}
    \bm{s}^{p,(i)} = \text{GELU}(\bm{W}^p\cdot \bm{h}^{i-1} + \bm{b}^p)
\end{equation}
and an utterance act network
\begin{equation}
    \bm{s}^{u,(i)} = \text{GELU}(\bm{W}^u\cdot \bm{d}^i + \bm{b}^u)
\end{equation}
where the probability of act $a_k$ is $p^{*,i}_k=exp(\bm{s}^{*,(i)}_k)$.
From these, we derive the policy act loss
\begin{equation}
    \mathcal{L}_p=\sum_k^{\left|\mathcal{A}\right|}y^a_{i,k}\log p^{p,i}_k + (1-y^a_{i,k})\log (1-p^{p,i}_k)
\end{equation}
and utterance act loss
\begin{equation}
    \mathcal{L}_u=\sum_k^{\left|\mathcal{A}\right|}y^a_{i,k}\log p^{u,i}_k + (1-y^a_{i,k})\log (1-p^{u,i}_k)
\end{equation}
for an utterance at $t=i$ with act labels $y^a_{i,k}$.

\paragraph{Like Prediction}
For every assistant message, the model predicts the likelihood of the user ``liking'' the message.
We treat this as binary classification, predict the ``like'' likelihood
\begin{equation}
    \hat{y}^l_i=\text{softmax}(\text{GELU}(\bm{W}^l\cdot \bm{h}^i + \bm{b}^l)),
\end{equation}
and use it to compute the like loss
\begin{equation}
    \mathcal{L}_l=\ell_{ce}(\hat{y}^l_i,y^l_i)
\end{equation}
where $y^l_i$ indicates if the message was liked.
We train the model jointly and optimize the loss
\begin{equation}
    \mathcal{L}=\mathcal{L}_f+\mathcal{L}_l+\mathcal{L}_p+\mathcal{L}_u.
\end{equation}
See Appendix~\ref{apx:method:train} for training details.

\section{Modeling Experiments}
\label{sec:exp}

\charm{} improves over a \bert{} model in most tasks.

\begin{table*}[t]
    \small
    \centering
    \IfFileExists{2020_emnlp_curiosity/commit_auto_fig/experiment-table.tex}{\begin{tabular}{l r r r r r r r r r r r r}
    \toprule                 & \multicolumn{2}{c}{Fact Rank (\textsc{mrr})} & \multicolumn{2}{c}{Utt. Act (\fone{})} & \multicolumn{2}{c}{Policy Act (\fone{})} & \multicolumn{2}{c}{Like (Accuracy)}                                                   \\
    \cmidrule(lr){2-3}\cmidrule(lr){4-5}\cmidrule(lr){6-7}\cmidrule(lr){8-9}
    Model                    & Val                                          & Test                                   & Val                                      & Test                                & Val   & Test           & Val   & Test           \\
    \midrule  Majority Class & \abr{n/a}                                    & \abr{n/a}                              & 0.602                                    & 0.604                               & 0.491 & 0.494          & 0.690 & 0.681          \\
    \abr{e2e} \bert{}        & 0.420                                        & 0.418                                  & 0.794                                    & 0.795                               & 0.635 & 0.631          & 0.829 & \textbf{0.822} \\
    \charm{}                 & 0.546                                        & \textbf{0.546}                         & 0.845                                    & \textbf{0.847}                      & 0.682 & \textbf{0.682} & 0.826 & 0.815          \\
    $-$ context              & 0.516                                        & 0.506                                  & 0.838                                    & 0.842                               & 0.664 & 0.664          & 0.824 & 0.820          \\
    \bottomrule\end{tabular}}{}
    \caption{
        The \charm{} model outperforms end-to-end \bert{} on most tasks.
        We compare fact selection with \mrr{}, dialog act prediction with micro-averaged \fone{}, and like prediction with accuracy.
        Ablating dialog history degrades context-dependent tasks (fact selection and policy act prediction), but not tasks more dependent on one message.
    }
    \label{tab:experiments}
\end{table*}

\subsection{Evaluation}
We evaluate each sub-task with separate metrics.
Fact selection is evaluated with mean reciprocal rank (\mrr{}).
For utterances with at least one selected fact, we compute the \mrr{} using the selected facts as relevant documents.
We compare like prediction with binary classification accuracy.
For utterance and policy act prediction, we compare models with micro-averaged \fone{} scores so that frequent classes are weighted more heavily.
For each metric, we report validation and test set scores.

\subsection{Baselines}
\bert{}~\citep{Devlin2018BERTPO} is a standard baseline for many \abr{nlp} tasks.
We use a multi-task extension of an uncased \bert{} model as our primary baseline and fine-tune it for our unique set of tasks~(\mtbert).
Specifically, we use the \abr{cls} representation of each utterance to replace the \hre{} representation as a time-distributed input to the same multi-task decoders~(Section \ref{subsec:method:task}).
The context-less \charm{} ablation replaces the dialog contextualizer \abr{lstm} with a per-timestep projection layer.
Lastly, we report majority class accuracy for classification tasks.

\subsection{Discussion}

The proposed \charm{} model for conversational curiosity is more effective than \mtbert{} for most of the tasks in \rover{} (Table~\ref{tab:experiments}).
Specifically, \charm{} improves significantly in fact prediction ($13$ \mrr{} points) and both dialog act prediction tasks ($5$ \fone{} points), demonstrating the efficacy of the structural encoding of the various input modalities.
Generally, models accurately predict utterance acts and likes, but their \mrr{} and \fone{} scores on fact selection and policy act prediction is comparatively worse.
To a degree, this is expected since there is not always one best fact or one best action to take as the assistant; there may be various reasonable choices, which is common in information retrieval tasks.
Nonetheless, models that specifically reason about the relationship between prior knowledge and entities would likely yield improvement.
For example, \citet{Liu2018KnowledgeDF} predict the most relevant unmentioned entity while~\citet{Lian2019LearningTS} model a posterior distribution over knowledge.
We leave these improvements to future work.

\section{Related Work}
\label{sec:rel}
Our work builds on knowledge-grounded conversational datasets and modeling.

\paragraph{Datasets} Although there are numerous grounded datasets, we did not find one for conversational information seeking that contained fine-grained knowledge groundings, message-level feedback from the user, and dialog acts.
Table~\ref{tab:datasets} compares the \rover{} dataset to several others according to six factors: (1) is the goal of the task information seeking, (2) is the dataset collected from natural dialog with one participant always taking the role of an assistant, (3) are dialog responses constrained, (4) are document groundings annotated---as opposed to distantly supervised---and fine-grained, (5) is there message level feedback for the assistant, and (6) is the dataset annotated with dialog acts.

\begin{table*}[t]
    \centering
    \small
    \begin{tabular}{p{5.3cm}cccccc}
        \toprule
        \multicolumn{1}{c}{Dataset}                                &
        \multicolumn{1}{p{1.15cm}}{\centering Info Seeking}        &
        \multicolumn{1}{p{1.5cm}}{\centering Dialog w/Assistant}   &
        \multicolumn{1}{p{1.3cm}}{\centering Free Response}        &
        \multicolumn{1}{p{1.45cm}}{\centering Annotated Grounding} &
        \multicolumn{1}{p{1.3cm}}{\centering Message Feedback}     &
        \multicolumn{1}{p{1cm}}{\centering Dialog Acts}                                                                              \\
        \midrule
        \rover{} (ours)                                            & \cmark & \cmark & \cmark    & \cmark    & \cmark    & \cmark    \\
        \midrule
        {Topical Chat}~\citep{Gopal2019topical}                    & \cmark & \dmark & \cmark    & \cmark    & \cmark    & \dmark    \\
        {Search as a Conversation}~\citep{ren2020search}           & \cmark & \cmark & \cmark    & \cmark    & \xmark    & \xmark    \\
        {Wizard of Wikipedia}~\citep{dinan2019wizard}              & \cmark & \cmark & \cmark    & \cmark    & \xmark    & \xmark    \\
        \quac{}~\citep{ChoiQuAC2018}                               & \cmark & \cmark & \xmark    & \cmark    & \xmark    & \dmark    \\
        \abr{cmu dog}~\citep{Zhou2018ADF}                          & \cmark & \cmark & \cmark    & \dmark    & \xmark    & \xmark    \\
        \abr{ms} {Marco Conv.}~\citep{Nguyen2016MSMA}              & \cmark & \xmark & \abr{n/a} & \abr{n/a} & \abr{n/a} & \abr{n/a} \\
        \opendialkg{}~\citep{moon-etal-2019-opendialkg}            & \xmark & \cmark & \cmark    & \cmark    & \xmark    & \xmark    \\
        \coqa{}~\citep{Reddy2018CoQAAC}                            & \xmark & \cmark & \dmark    & \cmark    & \xmark    & \xmark    \\
        {Holl-E}~\citep{Moghe2018TowardsEB}                        & \xmark & \dmark & \cmark    & \cmark    & \xmark    & \xmark    \\
        {Commonsense}~\citep{Zhou2018CommonsenseKA}                & \xmark & \xmark & \cmark    & \xmark    & \xmark    & \xmark    \\
        {Reddit+Wiki}~\citep{Qin2019ConversingBR}                  & \xmark & \xmark & \cmark    & \xmark    & \xmark    & \xmark    \\
        \bottomrule
    \end{tabular}
    \caption{
        \cmark~indicates a dataset has the feature,~\dmark~that it does with a caveat, and~\xmark~that it does not.
        Conversational \abr{ms marco} is a search dataset but has inquiry chains we want assistants to induce (exemplar in Appendix~\ref{apx:marco}).
        Topical Chat and Search as a Conversation are motivationally similar.
        While our dataset's combination of (human) annotation is unique, all three datasets are steps forward in resources for conversational information-seeking.
    }
    \label{tab:datasets}
\end{table*}

Our dataset is most similar to those for information-seeking such as \quac{}~\citep{ChoiQuAC2018}, Wizard of Wikipedia~\citep[\wow{}]{dinan2019wizard}, \abr{cmu dog}~\citep{Zhou2018ADF}, \abr{ms marco}~\citep{Nguyen2016MSMA}, Topical Chat~\citep{Gopal2019topical}, the \abr{trec} Conversational Assistance track~\citep[\cast{}]{Dalton2020TRECC2}, and Search as a Conversation~\citep[\saac{}]{ren2020search}.
\quac{} constrains assistant responses to spans from Wikipedia, which makes it better for conversational \emph{question answering}, but prevents more sophisticated assistant policies.
\quac{} also provides dialog acts, but they exist so that the assistant can inform the user of valid actions; we annotate dialog acts after-the-fact so that we can compare \emph{freely chosen} user responses.
Like \quac{}, Topical Chat, \saac{}, and \wow{} have annotated knowledge-groundings for each message, but responses are free-form.
\saac{} is a contemporaneous, \cast{}-based dataset that shares our motivation to make conversation a medium for information-seeking.
Topical Chat includes user feedback, but instead of explicitly defined roles, workers implicitly take dual and alternating roles as the user and assistant through knowledge asymmetry; followup work added automatically annotated dialog acts to Topical Chat~\citep{hedayatnia2020policy}.

Many tasks instruct annotators to take on a specific role in the dialog.
For example, in Wizard of Wikipedia, annotators assume an assigned persona~\citep{zhang-etal-2018-personalizing} in addition to being the user or assistant.
Consequently, many dialogs revolve around personal discussions instead of teaching about a topic.
Additionally, annotators may not have the background to play their role.
In contrast, we ask annotators to take roles that---as humans---they already know how to do: read about and convey interesting information on a topic (assistant) and engage in inquiry about a novel topic (user).

Our work is one of many in knowledge-grounded conversational datasets.
For example, \citet{Moghe2018TowardsEB} have workers discuss movies and ground messages to plot descriptions, reviews, comments, and factoids; however, one worker plays both roles.
In \opendialkg{}~\cite{moon-etal-2019-opendialkg}, annotators ground messages by path-finding through Freebase~\citep{Bast2014EasyAT} while discussing and recommending movies, books, sports, and music.
\citet{Qin2019ConversingBR} use Reddit discussion threads as conversations and ground to web pages.
Similarly, \citet{ghazvininejad2018a} collect Twitter three-turn threads and ground to Foursquare restaurant reviews.
Our work adds to this dataset compendium.

\paragraph{External Knowledge in Models} Our model is related to those that incorporate external information like facts in question answering~\citep{Weston2015MemoryN,Sukhbaatar2015EndToEndMN,kvnets2016}, knowledge base triples in dialog models~\citep{Han2015ExploitingKB,He2017LearningSC,Parthasarathi2018ExtendingNG}, common sense~\citep{Young2017AugmentingED,Zhou2018CommonsenseKA}, or task-specific knowledge~\citep{Eric2017KeyValueRN}.
Similarly to~\citet{kalchbrenner-blunsom-2013-recurrent,khanpour-etal-2016-dialogue}, \charm{} predicts the act of the current message, but also next message's act like~\citet{tanaka-etal-2019-dialogue} do.

\section{Future Work and Conclusion}
\label{sec:fw}

We see two immediate directions for future work.
The first is to augment our \charm{} model with a text generation module to make a digital version of our human assistants.
This involves contextualizing and paraphrasing facts which our dataset supports.
Second, dialog act sequences could identify additional data-driven policies that could be used to define rewards or losses.
By conditioning on dialog acts or sequences of dialog acts, textual outputs could be better-controlled~\citep{Sankar2019DeepRL,See2019WhatMA} and combined with knowledge grounding~\citep{hedayatnia2020policy}.
However, text is not the native modality of digital assistants.

We envision digital assistants participating in information-seeking, which means handling speech input.
Consequently, automatic speech recognition (\abr{asr}) introduces transcription errors which are especially prevalent in knowledge-oriented text like question answering~\citep{peskov2019noisy}.
\citet{gopalakrishnan2020asr} show this is also problematic in information-seeking dialog by comparing models on textual and \abr{asr} versions of Topical Chat.
To close the loop in conversational information-seeking, models need to account for the speech-based environment of digital assistants.

In summary, this work introduces \rover{}: a large-scale conversational information seeking dataset.
With \rover{}'s unique set of annotations, we design \charm{} which jointly learns to choose facts, predict a policy for the next message, classify dialog acts of messages, and predict if a message will be liked.
We hope that our dataset will encourage further interest in curiosity-driven dialog.

\section*{Acknowledgments}

We thank Rajen Subba, Stephen Roller, Alborz Geramifard, and Scott Yih for insightful discussions.
Thanks to Becka Silvert for improving the task guidelines and Victor Ling for building the dialog act annotation tool's backend.
Thanks to Sara LaRocca and Hal Daum\'e III for advice on adapting Krippendorff's $\alpha$.
We thank our anonymous \abr{acl} and \abr{emnlp} reviewers as well as Shi Feng, Jordan Boyd-Graber, Joe Barrow, Karthik Gopalakrishnan, and \abr{umd} \abr{clip} members for feedback on the paper.
Rodriguez's contributions to this work while at \abr{umd} were supported by \abr{nsf} Grant \abr{iis}-1822494.
Any opinions, findings, conclusions, or recommendations expressed here are those of the authors and do not necessarily reflect the view of the sponsor.

\bibliography{bib/journal-full,bib/pedro}
\bibliographystyle{style/acl_natbib}
\clearpage
\begin{appendix}
  \appendix

\section{Components of Dialog Interfaces}
\label{apx:int-photos}
In this section, we provide short descriptions and screenshots of every component of the user and assistant dialog interfaces.

\subsection{User's Interface}

\begin{figure}[ht]
    \centering
    \nicebox{
        \includegraphics[width=\linewidth]{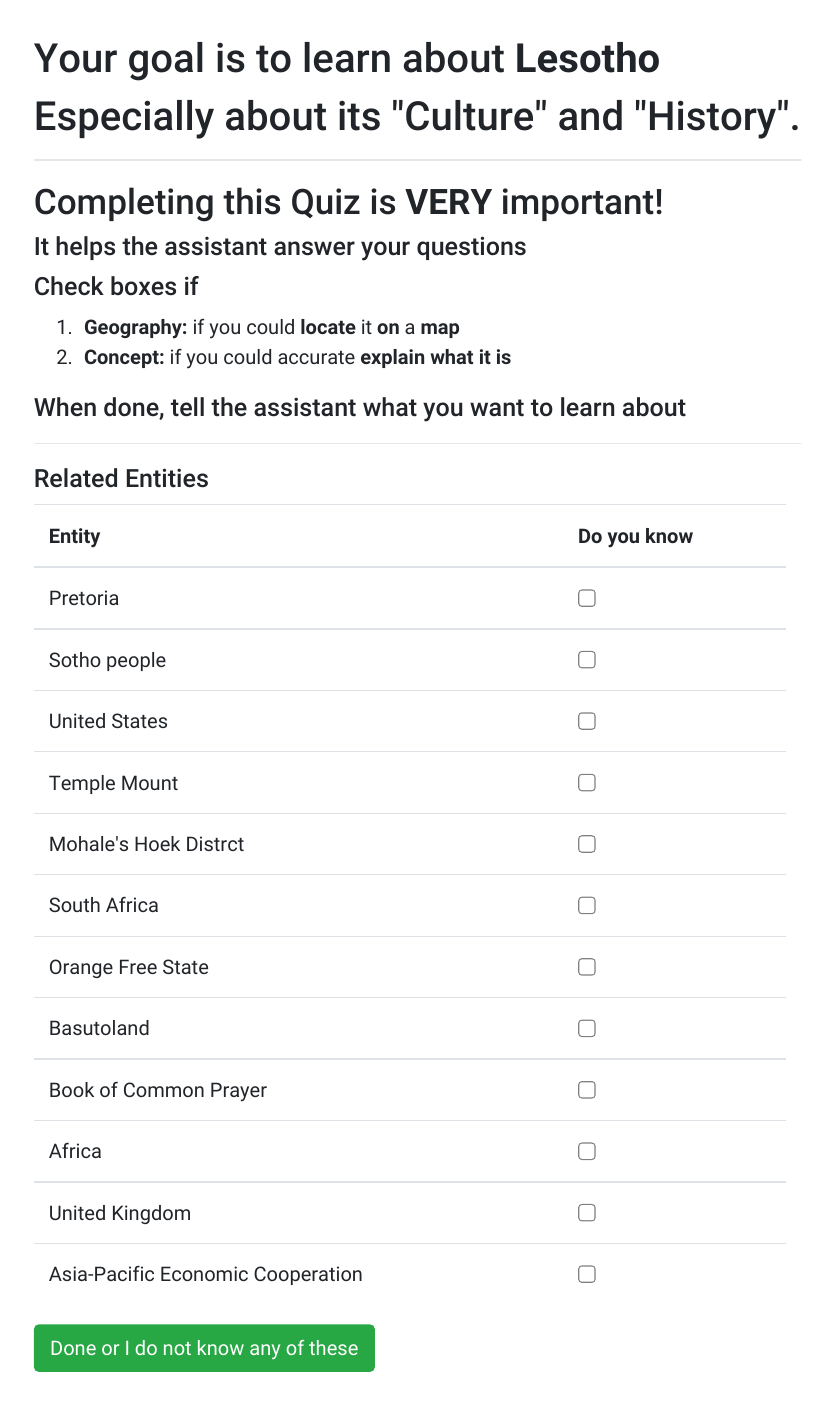}
    }
    \caption{
        In this example, the user is assigned to learn about \topic{Lesotho}, specifically its \aspect{culture} and \aspect{history}.
        In addition to their training with guidelines and videos, we repeat the instructions here.
        The related entities span relatively common ones like the \entity{United States} or \entity{Africa} to less known ones such as \entity{Basutoland}.
    }
    \label{fig:entity-quiz}
\end{figure}

Figure~\ref{fig:entity-quiz} shows the interface that we use to sample the user's prior knowledge of entities related to the topic.
To derive a diverse sample, we use Wikipedia page views as a proxy for how well known the entity is.
All experiments use the English Wikipedia dump generated on July 23, 2019.
We divide entity mentions into ten buckets based on the frequency of page views, and round-robin sample fifteen entities from those buckets.
The interface is shown before the user starts chatting with the assistant.

\begin{figure*}[ht]
    \centering
    \nicebox{
        \begin{minipage}{.35\textwidth}
            \includegraphics[width=\linewidth]{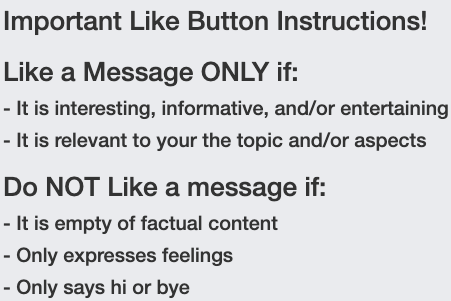}
        \end{minipage}
        \begin{minipage}{.60\textwidth}
            \includegraphics[width=\linewidth]{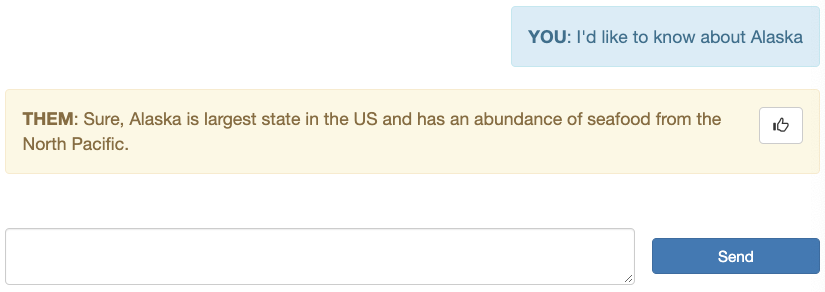}
        \end{minipage}
    }
    \caption{
        The user expresses the ``interestingness'' of the assistant's messages through a ``like'' button (right of message).
        The instructions are shown prominently in the full interface and repeated in training material.
    }
    \label{fig:like-button}
\end{figure*}

We elicit how ``interesting'' a user finds each of the assistant's messages through the like button in Figure~\ref{fig:like-button}.
Only users can ``like'' a message; the assistant cannot ``like'' user messages.
Users are instructed to ``like'' messages if they are ``interesting, informative and/or entertaining'' and ``relevant to their topic and/or aspects.''
They are specifically instructed not to ``like'' messages that are devoid of factual content, only express feelings, or only contain greetings or farewells.

\begin{figure}[ht]
    \centering
    \nicebox{\includegraphics[width=.98\linewidth]{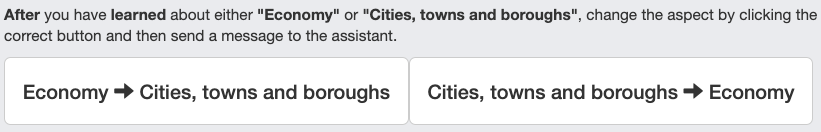}}
    \caption{
        The user is assigned two aspects about their topic.
        After they are satisfied with what they have learned about the first aspect, they click a button and switch to the next aspect.
        While the button click is not communicated to the assistant (the user must send a corresponding message), it resets the fact contextualizer; we observe that without this, too many facts were related to the previous aspect.
    }
    \label{fig:switch-aspect}
\end{figure}

\paragraph{Switching Aspect}
Users are randomly assigned two aspects for each dialog and told to spend time discussing each.
The guidelines instruct them to spend at least two turns per topic, but we do not specify any further time requirements.
When the user changes aspects, we instruct them to click a button (Figure~\ref{fig:switch-aspect}) to indicate when and which aspect they are switching to.
Additionally, this event triggers a reset in the context we use to rank the assistant's facts.

\subsection{Assistant Interface}

By design, we intend for most workers to not be familiar in depth with most of the geographic topics.
Thus, the most important responsibility of the assistant interface is to provide enough information---without overwhelming them---to be engaging conversational partners.
The first interface shown is a short description of the topic from either Simple Wikipedia or the English Wikipedia.
This component helps the assistant reach a general understanding of the topic so that they can choose better facts.
\begin{figure}[ht]
    \centering
    \nicebox{\includegraphics[width=.98\linewidth]{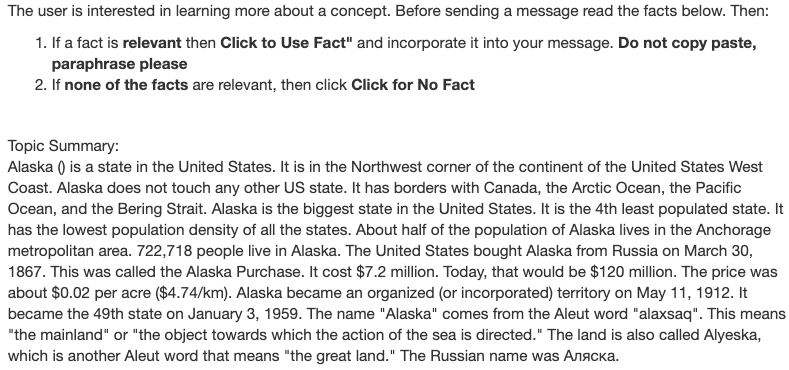}}
    \caption{
        A short topic description is always visible to the assistant.
        The goal is to ensure the assistant always has a general understanding of the dialog topic.
    }
    \label{fig:summary}
\end{figure}

\begin{figure*}[ht]
    \centering
    \begin{minipage}{\textwidth}
        \includegraphics[width=\linewidth]{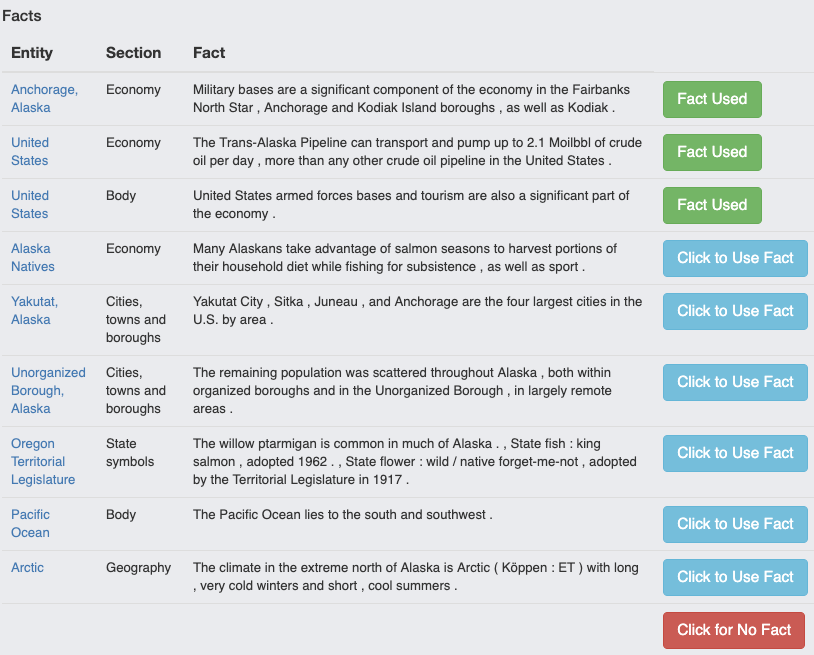}
    \end{minipage}
    \begin{minipage}{\textwidth}
        \nicebox{
            \includegraphics[width=.98\linewidth]{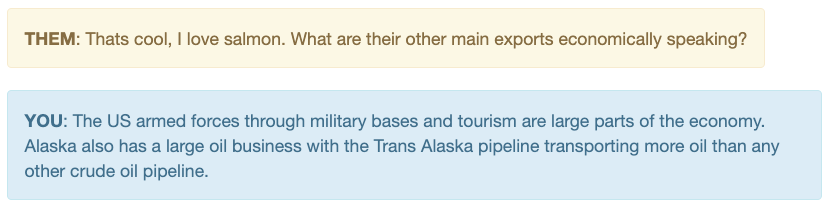}
        }
    \end{minipage}

    \caption{
        The assistant could incorporate any number of facts into their reply to the user.
        Their goal was to answer the user's immediate questions, and anticipate what information they would be most interested in.
    }
    \label{fig:grounded-msg}
\end{figure*}

The most important component of the assistant interface is their list of available facts.
These facts have high textual similarity with the most recent three turns and are broken into three categories: facts related to entities the user knows about (rooted facts), facts related to an aspect (aspect facts), and facts from anywhere on the page (general facts).
Feedback from pilot collections showed that six facts was too few which caused a lack of relevant facts, but twelve facts overwhelmed annotators.
Thus, we use nine facts so that we can also balance equally across each type of fact.
When composing their reply, the assistant can use any number of facts as in Figure~\ref{fig:grounded-msg}.
To discourage verbatim copying, we disable the paste feature in javascript.
We also drop repeatedly unused facts.

\section{Dialog Act Annotation}
\label{apx:acts}
\begin{figure*}[ht]
    \centering
    \includegraphics[width=\linewidth]{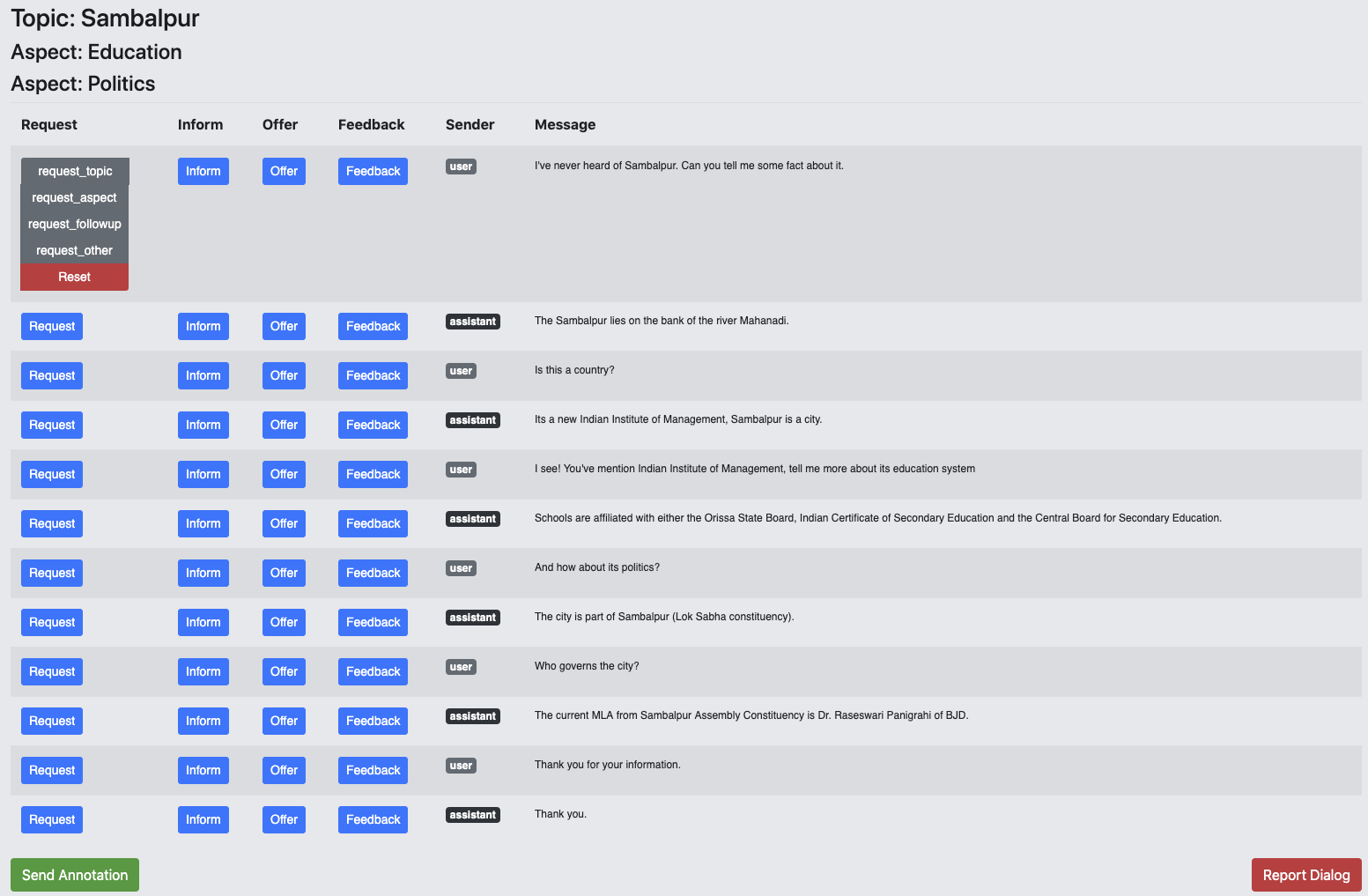}
    \caption{
        To annotate dialog acts, we develop an interface that showed each utterance on a separate line.
        Annotators assign zero or more dialog acts to each utterance using grouped dropdowns.
    }
    \label{fig:da-iface}
\end{figure*}

To annotate dialog acts, we create a separate annotation interface (Figure~\ref{fig:da-iface}).
The interface shows one dialog at a time, and the same annotator annotates all the utterances.
In addition to the utterances, the interface shows the topic, aspects, and sender of each message.
Lastly, we incorporate a ``Report Dialog'' feature to help identify and remove inappropriate dialogs.

\section{Sample Dialogs}
\label{apx:samples}
Tables~\ref{tab:dialog_example1} and~\ref{tab:dialog_example2} show \rover{} dialogs and highlight the dataset's features.
Typos and grammatical errors made by annotators are left unaltered.

\begin{table*}[t]
    \small
    \begin{center}
        \setlength\tabcolsep{3.5pt}
        \begin{tabular}{p{0.2cm}p{3.8cm}p{11cm}}
            \toprule[\heavyrulewidth]
              & \multicolumn{1}{r}{\textbf{Topic}:}          & \underline{Djibouti}                                                                                                                                                                                                                                                                                      \\
              & \multicolumn{1}{r}{\textbf{Aspects}:}        & \textit{Economy}, \textit{Politics}                                                                                                                                                                                                                                                                       \\
              & \multicolumn{1}{r}{\textbf{Known Entities}:} & France, Africa                                                                                                                                                                                                                                                                                            \\
            \midrule
              & Dialog Acts                                  & Messages                                                                                                                                                                                                                                                                                                  \\
            \midrule
            U & \texttt{request topic}                       & Hi can you tell me about Djibouti?                                                                                                                                                                                                                                                                        \\
            A & \texttt{inform response}                     & The country of Djibouti is officially called the Republic of Djibouti is located in the horn of Africa. It was created out of French Somaliland. It is also recorded the country was colonized by France in1800s and later gained its independence on June 27, 1977.                                      \\
            U & \texttt{request aspect}                      & Can you tell me about their Politics                                                                                                                                                                                                                                                                      \\
            A & \texttt{inform response}                     & After its independence, The country had two regiments commanded by the French officers. The country also maintains its close ties with Somalia, Ethiopia, France and the United States.                                                                                                                   \\
            U & \texttt{request aspect}                      & Any other information about the French and their impact of the politics?                                                                                                                                                                                                                                  \\
            A & \texttt{inform response}                     & There are no direct information on its effect on the country's politics but even though the country is already independent. The 13th Demi-brigade of the French Foreign Legion is still stationed in this country. It is also the largest french military presence abroad. commanded by a 3-star general. \\
            U & \texttt{request aspect}                      & Heavy French infeleunce. what other interesting inofrmation on politics do you have?                                                                                                                                                                                                                      \\
            A & \texttt{inform response}                     & In recent years, the country has improved its training techniques, military command, and information structure. They also became more self-reliant in supplying their military and assist in peacekeeping missions and provide military assistance to countries that officially ask them for help.        \\
            U & \texttt{request aspect, feedback positive}   & Wow Interesting, Can you now tell me about the economy they have?                                                                                                                                                                                                                                         \\
            A & \texttt{offer aspect}                        & Sorry, I do not have that information. But I can tell you more about Politics.                                                                                                                                                                                                                            \\
            U & \texttt{offer accept}                        & Sure                                                                                                                                                                                                                                                                                                      \\
            A & \texttt{inform response}                     & Camp Lemonnier is rented by the United State as a Naval Expeditionary Based for \$63 Million a year. While Japan and France each pays \$30 million a year and China pays \$20 Million a year.                                                                                                             \\
            U & \texttt{request other}                       & Lastly, any any fun facts?                                                                                                                                                                                                                                                                                \\
            A & \texttt{inform response}                     & I am not sure if this is a fun fact, but the country's gross domestic product expanded by more than 6\%. From \$341 Million to 1.5 Billion                                                                                                                                                                \\
            U & \texttt{}                                    & That's a huge increase. thank you for all your help                                                                                                                                                                                                                                                       \\
            A & \texttt{}                                    & You are welcome                                                                                                                                                                                                                                                                                           \\
            \bottomrule[\heavyrulewidth]
        \end{tabular}
    \end{center}
    \caption{Example dialog \#1 from \rover{}. (U: User, A: Assistant)}
    \label{tab:dialog_example1}
\end{table*}

\begin{table*}[t]
    \small
    \begin{center}
        \setlength\tabcolsep{3.5pt}
        \begin{tabular}{p{0.2cm}p{3.8cm}p{11cm}}
            \toprule[\heavyrulewidth]
              & \multicolumn{1}{r}{\textbf{Topic}:}          & \underline{British Columbia}                                                                                                                                                                         \\
              & \multicolumn{1}{r}{\textbf{Aspects}:}        & \textit{Government and politics}, \textit{Culture}                                                                                                                                                   \\
              & \multicolumn{1}{r}{\textbf{Known Entities}:} & Canada, Seattle                                                                                                                                                                                      \\
            \midrule
              & Dialog Acts                                  & Messages                                                                                                                                                                                             \\
            \midrule
            U & \texttt{request topic}                       & Hi! Can you help me learn some basic information about British Columbia? I don't know much except that it's located in Canada.                                                                       \\
            A & \texttt{inform response}                     & Yes, British Columbia is the westernmost province of Canada and is located between the Rocky Mountains and the Pacific Ocean.                                                                        \\
            U & \texttt{request aspect, feedback positive}   & I didn't know it was on the coast! What can you tell me about government and politics there?                                                                                                         \\
            A & \texttt{inform response}                     & One interesting fact about the government is that the Green Part plays a larger role in this province than it does in other provinces of Canada.                                                     \\
            U & \texttt{request followup, feedback positive} & Interesting. What can else you tell me about the Green Party?                                                                                                                                        \\
            A & \texttt{inform response}                     & The New Democratic Party and the Green Party caucuses together control 44 seats. Which seems like a lot but the British Columbia Green Party only takes up 3 of those 44 seats.                      \\
            U & \texttt{request aspect}                      & That's a pretty small influence. Can you tell me some fun culture facts about British Columbia?                                                                                                      \\
            A & \texttt{}                                    & I am sorry I do not have any information on their culture right now.                                                                                                                                 \\
            U & \texttt{request topic}                       & That's okay. What other fun facts can you share?                                                                                                                                                     \\
            A & \texttt{inform response}                     & Interestingly, Queen Victoria chose British Columbia to distinguish what was the British sector of the Columbia District from the United States which became the Oregon Territory on August 8, 1848. \\
            U & \texttt{request aspect}                      & So that's why it has "British" specifically as part of it's name! Makes sense. Are there any sports or outdoor activities that are popular in British Columbia?                                      \\
            A & \texttt{inform response}                     & Horseback riding is enjoyed by many British Columbians.                                                                                                                                              \\
            U & \texttt{}                                    & Thanks for your help today. Now I know more than I did before.                                                                                                                                       \\
            A & \texttt{}                                    & No problem, it was a pleasure.                                                                                                                                                                       \\
            \bottomrule[\heavyrulewidth]
        \end{tabular}
    \end{center}
    \caption{
        Example dialog \#2 from \rover{}. (U: User, A: Assistant).
        After mentioning the Green Party, the user asks a specific followup question; we use these interactions to estimate implicit preference.
    }
    \label{tab:dialog_example2}
\end{table*}

\section{Paraphrase Analysis and Samples}
\label{apx:para}
In Section~\ref{sec:para-analysis}, we describe the results of a manual analysis on two hundred and fifty assistant paraphrases.
Annotations were completed by the authors and shown in Table~\ref{tab:para}.
We break messages into four categories: paraphrases, copies, errors, and unrelated.
Paraphrases include messages that incorporate the selected fact and possibly additional information.
Copies include verbatim copying, cherry-picked phrases, and trivial contextualizations like replacing an entity with a pronoun.
Table~\ref{tab:par-ex} shows ten randomly selected paraphrases from the two hundred and fifty manual annotations.

\begin{table*}
    \small
    \centering
    \IfFileExists{2020_emnlp_curiosity/commit_auto_fig/paraphrase-table.tex}{\begin{tabular}{l l r r}
    \toprule Category & Label & Count & Percent \\ \midrule  Copy & verbatim & 68 & $27.2\%$\\  Copy & cherry-pick & 6 & $2.40\%$\\  Copy & context & 30 & $12.0\%$\\  \midrule Copy & Total & 104 & $41.6\%$\\\midrule\\  Paraphrase & paraphrase-correct & 111 & $44.4\%$\\  Paraphrase & paraphrase-multiple & 17 & $6.80\%$\\  \midrule Paraphrase & Total & 128 & $51.2\%$\\\midrule\\  Error & paraphrase-error & 5 & $2.00\%$\\  Unrelated & unrelated & 13 & $5.20\%$\\  \bottomrule Total && 250 & $100\%$\\  \bottomrule\end{tabular}}{}
    \caption{
        We analyze the paraphrases annotators use through manual categorization.
        The ``Copy'' category includes cherry-picked verbatim phrases, verbatim copies, and contextualized copies (e.g., changing a named entity to ``it'').
        The majority of paraphrases are correct and only incorporate the provided fact, but a few weave in other information.
        $7.2\%$ of paraphrases are either unrelated to the selected facts or paraphrase the fact incorrectly.
        Overall, $51.2\%$ of messages have valid paraphrases.
    }
    \label{tab:para}
\end{table*}

\begin{table*}
    \small
    \centering
    \begin{tabular}{p{15cm}}
        \toprule
        \textbf{Label:} paraphrase-correct                                                                                                                                                                                                                                                                                                                                                                           \\
        \textbf{Message:} I do not have the information available but I can tell you 10 million tones were exported annually for, Cardiff dock in the years leading up to the First World War.                                                                                                                                                                                                                       \\
        \textbf{Fact:} In the years leading up to the First World War , more than 10 million tonnes of coal was exported annually from Cardiff Docks .                                                                                                                                                                                                                                                               \\
        \midrule
        \textbf{Label:} copy                                                                                                                                                                                                                                                                                                                                                                                         \\
        \textbf{Message:} Media and journalists in Namibia are represented by the Namibia chapter of the Media Institute of 'southern Africa and the Editors Forum of Namibia.                                                                                                                                                                                                                                       \\
        \textbf{Fact:} Media and journalists in Namibia are represented by the Namibian chapter of the Media Institute of Southern Africa and the Editors ' Forum of Namibia .                                                                                                                                                                                                                                       \\
        \midrule
        \textbf{Label:} paraphrase-multiple                                                                                                                                                                                                                                                                                                                                                                          \\
        \textbf{Message:} The film industry has a rich history in Indonesian culture and the Indonesian Film Festival has been held regularly since 1955.                                                                                                                                                                                                                                                            \\
        \textbf{Fact:} Indonesia has held annual film festivals and awards , including the Indonesian Film Festival ( Festival Film Indonesia ) that has been held intermittently since 1955 .                                                                                                                                                                                                                       \\
        \midrule
        \textbf{Label:} context                                                                                                                                                                                                                                                                                                                                                                                      \\
        \textbf{Message:} EU means European Union. GDP per capital ranges from 188\% of EU average in Bratislava to 54\% in eastern Slovakia                                                                                                                                                                                                                                                                         \\
        \textbf{Fact:} GDP per capita ranges from 188 \% of EU average in Bratislava to 54 \% in Eastern Slovakia .                                                                                                                                                                                                                                                                                                  \\
        \midrule
        \textbf{Label:} paraphrase-correct                                                                                                                                                                                                                                                                                                                                                                           \\
        \textbf{Message:} Sudan was considered the 17th growing economy in the world. Its faster development of the country from oil profits.                                                                                                                                                                                                                                                                        \\
        \textbf{Fact:} Sudan , officially the Republic of the Sudan , is a country in Northeast Africa .                                                                                                                                                                                                                                                                                                             \\
        \midrule
        \textbf{Label:} paraphrase-correct                                                                                                                                                                                                                                                                                                                                                                           \\
        \textbf{Message:} There is a mountain range called the Annamite Range that runs along eastern border to Vietnam and another range, named Luang Prabang Range, near the Thailand border.                                                                                                                                                                                                                      \\
        \textbf{Fact:} The Mekong River forms a large part of the western boundary with Thailand , where the mountains of the Annamite Range form most of the eastern border with Vietnam and the Luang Prabang Range the northwestern border with the Thai highlands .                                                                                                                                              \\
        \midrule
        \textbf{Label:} copy                                                                                                                                                                                                                                                                                                                                                                                         \\
        \textbf{Message:} In 1603, the kingdoms of England, Scotland and Ireland were united in a personal union when James VI, king of Scots, inherited the crowns of England and Ireland and moved his court from Edinburgh to London; each country nevertheless remained a separate political entity and retained its separate political, legal and religious institutions. Would you like to know anything else? \\
        \textbf{Fact:} In 1603 , the kingdoms of England , Scotland and Ireland were united in a personal union when James VI , King of Scots , inherited the crowns of England and Ireland and moved his court from Edinburgh to London ; each country nevertheless remained a separate political entity and retained its separate political , legal , and religious institutions .                                 \\
        \midrule
        \textbf{Label:} copy                                                                                                                                                                                                                                                                                                                                                                                         \\
        \textbf{Message:} The A31 connects to the A35 at Bere Regis , and passes east through Wimborne and Ferndown to Hampshire where it later becomes the M27                                                                                                                                                                                                                                                      \\
        \textbf{Fact:} The A31 connects to the A35 at Bere Regis , and passes east through Wimborne and Ferndown to Hampshire , where it later becomes the M27 .                                                                                                                                                                                                                                                     \\
        \midrule
        \textbf{Label:} context                                                                                                                                                                                                                                                                                                                                                                                      \\
        \textbf{Message:} I'm not sure of that. One fact I do have is that by 1857, Chicago was Illinois' largest city.                                                                                                                                                                                                                                                                                              \\
        \textbf{Fact:} By 1857 , Chicago was Illinois 's largest city .                                                                                                                                                                                                                                                                                                                                              \\
        \midrule
        \textbf{Label:} paraphrase-correct                                                                                                                                                                                                                                                                                                                                                                           \\
        \textbf{Message:} On March 2, 2008, Dmitry Medvedev was elected President of Russia and Vladimir Putin became Prime Minister.                                                                                                                                                                                                                                                                                \\
        \textbf{Fact:} On 2 March 2008 , Dmitry Medvedev was elected President of Russia while Putin became Prime Minister .                                                                                                                                                                                                                                                                                         \\
        \bottomrule
    \end{tabular}
    \caption{
        A random sample of ten manually labeled paraphrases from the assistant.
        The top row indicates the label we (the authors) annotated, the middle row the message, and the bottom row the original fact from Wikipedia.
        The original fact is shown as displayed to crowd-workers including punctuation tokenization.
    }
    \label{tab:par-ex}
\end{table*}

\section{Like Prediction Comparison}
\label{apx:like-compare}
Like prediction is the one task where \charm{} was not the best model.
To better understand the differences between the \charm{} and \bert{} model, we randomly sample thirty dialogs in the test set and find assistant messages where the model predictions disagree.
Of the 202 assistant messages in the thirty dialogs, the \bert{} like prediction is correct 174 times ($86.1\%$) and \charm{} 170 times ($84.2\%$).
The models disagree on eight predictions ($3.96\%$); \bert{} is correct on six of these and \charm{} on the other two.
Table~\ref{tab:like-compare} shows the messages, predictions, and labels.
Although we might guess at why the predictions differ (e.g., perhaps something in the Florida dialog confused \bert{}), without further analysis we cannot establish causality.

\begin{table*}[ht]
    \small
    \centering
    \begin{tabular}{l l p{12cm}}
        \toprule
        Liked & Correct Model & Message                                                                                                                                                                          \\
        \midrule
        No    & \bert{}       & You are welcome!                                                                                                                                                                 \\
        \midrule
        Yes   & \bert{}       & I'm sorry I don't have anymore information about the etymology of Tunisia, but what I can tell you is that Tunisia Sports City is a whole sports city being constructed in Tunis \\
        \midrule
        Yes   & \bert{}       & Yes Buddhism is a dominant influence in Lao culture. It has been great helping you.                                                                                              \\
        \midrule
        Yes   & \charm{}      & Florida is a state in the southeast United States. What would you like to know?                                                                                                  \\
        Yes   & \bert{}       & They have an average daily temperature of 70.7, it's the warmest state in the U. S.                                                                                              \\
        Yes   & \charm{}      & Yes, I can. Florida is nicknamed the ``Sunshine State'', but severe weather is a common occurrence.                                                                              \\
        \midrule
        Yes   & \bert{}       & Hello, Indonesia is part of the Malay Islands and is in Southeast Asia. Would you like to know more about the history?                                                           \\
        Yes   & \bert{}       & I do not have etymologic information, would you like to know more about the economy? I can tell you thank Indonesia develops military and commuter aircraft.                     \\
        \bottomrule
    \end{tabular}
    \caption{
        To compare like prediction between models, we randomly sample thirty dialogs and obtain predictions from \charm{} and \bert{}.
        The table only shows messages where the model predictions disagree and indicates which model was correct.
        Dialogs are delineated by horizontal lines.
        Unfortunately, from only these examples we cannot determine why the \charm{} model errors in most of these predictions.
    }
    \label{tab:like-compare}
\end{table*}

\section{Model Training, Implementation, and Computation}
\label{apx:method:train}

We implement all models with PyTorch~\citep{paszke2017automatic} and \allennlp{}~\citep{Gardner2018AllenNLPAD}.
The learning rates for models is set using the built-in learning rate finder in \allennlp{}.
Model losses were optimized with Adam~\citep{Kingma2014AdamAM}; the \bert{} model uses a learning rate of $.0001$ and \charm{} a learning rate of $.001$ with otherwise default parameters.
We train for a maximum of forty epochs and early stop based on the sum of validation losses.
The \charm{} model uses batch size $64$ and the \bert{} model batch size $4$.
Our best model (\charm{}), has $26,970,475$ parameters, takes two hours and eighteen minutes to train, and early stops on epoch fifteen.
In our models, text encoders for utterances and facts share parameters.

Models were developed on a single machine with eighty Intel $2.0$GHz \abr{cpu}s, $256$\abr{gb} \abr{ram}, and eight Tesla V100 graphics cards.
Each model was trained and evaluated on a single graphics cards with hyper-parameter sweeps parallelized across the eight cards.

\allennlp{} configuration files and software dependencies (including version) are included in our code at \href{https://github.com/facebookresearch/curiosity}{github.com/facebookresearch/curiosity}.

\section{MS Marco Conversational Sample Queries}
\label{apx:marco}

Conversational \abr{ms marco} is a search dataset that partially inspired this work.
Assistant messages should prompt followup queries like in Table~\ref{tab:marco}.

\begin{table*}[t]
    \small
    \begin{center}
        \begin{tabular}{l}
            \toprule
            Query                                                                             \\
            \midrule
            What is a physician's assistant?                                                  \\
            What are the educational requirements required to become a physician's assistant? \\
            What does the education to become a physician's assistant cost?                   \\
            What's the average starting salary of a physician's assistant in the UK?          \\
            What's the average starting salary of a physician's assistant in the US?          \\
            What school subjects are needed to become a registered nurse?                     \\
            What is the physician's assistant average salary vs a registered nurse?           \\
            What the difference between a physician's assistant and a nurse practitioner?     \\
            Do nurse practitioners or physician's assistant's make more?                      \\
            Is a physician's assistant above a nurse practitioner?                            \\
            What is the fastest way to become a nurse practioner?                             \\
            How much longer does it take to become a doctor after being a nurse practitioner? \\
            What are the main breeds of goat?                                                 \\
            Tell me about boer goats.                                                         \\
            What goat breed is good for meat?                                                 \\
            Are angora goats good for meat?                                                   \\
            Are boer goats good for meat?                                                     \\
            What are pygmy goats used for?                                                    \\
            What goat breed is the best for fiber production?                                 \\
            How long do Angora goats live?                                                    \\
            Can you milk Angora goats?                                                        \\
            How many Angora goats can you have per acre?                                      \\
            Are Angora goats profitable?                                                      \\
            \bottomrule
        \end{tabular}
    \end{center}
    \caption{
        An exemplar query chain from the conversational variant of \abr{ms marco}.
        An ideal assistant should answer these questions \emph{and} inspire these types of followup questions.
    }
    \label{tab:marco}
\end{table*}

\end{appendix}
\end{document}